%% file: main.tex
\documentclass[10pt,twocolumn,letterpaper]{article}

\usepackage[pagenumbers]{iccv} % To force page numbers, e.g. for an arXiv version

\usepackage{colortbl}


\definecolor{iccvblue}{rgb}{0.21,0.49,0.74}
\usepackage[pagebackref,breaklinks,colorlinks,allcolors=iccvblue]{hyperref}

\title{ContextFormer: Redefining Efficiency in Semantic Segmentation}

\author{Mian Muhammad Naeem Abid, Nancy Mehta, Zongwei Wu, Radu Timofte\\
Computer Vision Lab, CAIDAS, University of Würzburg, Germany\\
}

\begin{document}
\maketitle
\input{sec/0_abstract}    
\input{sec/1_intro}

\input{sec/2_Related_Work}
\input{sec/3_Proposed_Method}
\input{sec/4_Experiments}
\input{sec/5_Conclusion}

{
    \small
    \bibliographystyle{ieeenat_fullname}
    \bibliography{main}
}
\input{sec/X_suppl}

\end{document}

%% file: sec/0_abstract.tex
\begin{abstract}
Semantic segmentation assigns labels to pixels in images, a critical yet challenging task in computer vision. Convolutional methods, although capturing local dependencies well, struggle with long-range relationships. Vision Transformers (ViTs) excel in global context capture but are hindered by high computational demands, especially for high-resolution inputs. Most research optimizes the encoder architecture, leaving the bottleneck underexplored—a key area for enhancing performance and efficiency. We propose ContextFormer, a hybrid framework leveraging the strengths of CNNs and ViTs in the bottleneck to balance efficiency, accuracy, and robustness for real-time semantic segmentation. The framework's efficiency is driven by three synergistic modules: the Token Pyramid Extraction Module (TPEM) for hierarchical multi-scale representation, the \textbf{Trans}former and \textbf{B}ranched \textbf{D}epthwise\textbf{C}onv (Trans-BDC) block for dynamic scale-aware feature modeling, and the Feature Merging Module (FMM) for robust integration with enhanced spatial and contextual consistency.
Extensive experiments on ADE20K, Pascal Context, CityScapes, and COCO-Stuff datasets show ContextFormer significantly outperforms existing models, achieving state-of-the-art mIoU scores, setting a new benchmark for efficiency and performance. The codes will be made publicly available upon acceptance.
\end{abstract}

%% file: sec/1_Intro.tex
\section{Introduction}
\label{sec:intro}

\begin{figure}[t]
  \centering
  \includegraphics[width=\linewidth]{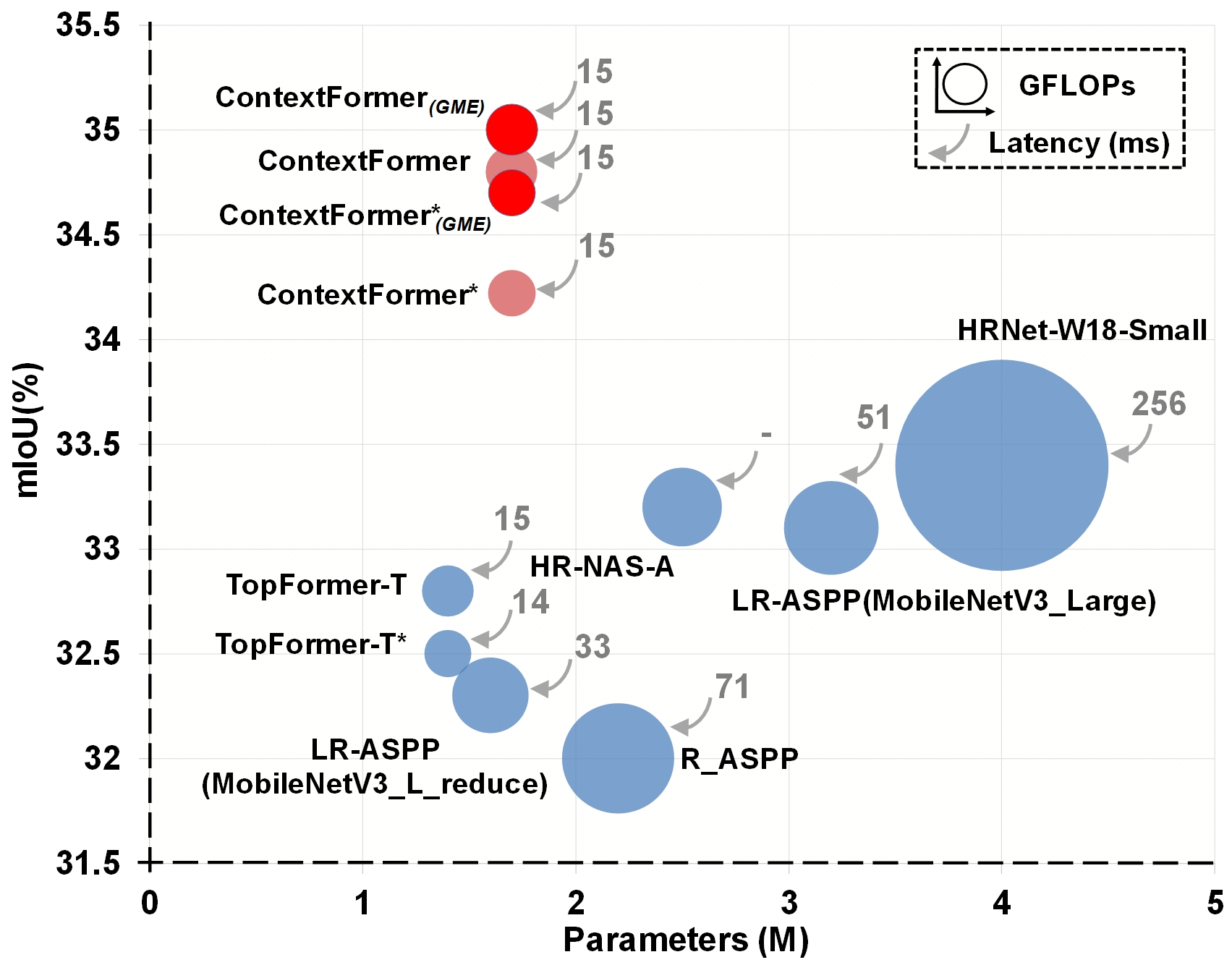}
  \caption{Comparison of mIoU, parameters, GFLOPs, and latency for state-of-the-art efficient models on the ADE20K validation set. Our ContextFormer strikes an optimal balance between performance and efficiency, delivering lower latency while maintaining competitive results. The circle sizes represent GFLOPs, with * indicating models trained on $448 \times 448$ input resolution.}
  \label{fig:comparison_of_all_models}
\end{figure}

Semantic segmentation is a fundamental task in computer vision that involves dense pixel-wise classification~\cite{mo2022review}. Traditional methods~\cite{he2016deep, he2017p, ronneberger2015u, chen2018deeplabv3+, badrinarayanan2017segnet} rely on Convolutional Neural Networks (CNNs) to extract hierarchical feature representations using local receptive fields. However, convolutional operators inherently struggle with capturing long-range dependencies, leading to suboptimal segmentation accuracy in complex scenes. With the advent of Vision Transformers (ViTs) \cite{dosovitskiy2020image}, segmentation models have demonstrated superior global context modeling \cite{cheng2022masked, xie2021segformer}. Despite their effectiveness, ViTs exhibit quadratic complexity in self-attention computations, leading to excessive memory and compute requirements for high-resolution inputs~\cite{dehghani2023scaling, liu2022swin, li2022efficientformer, mehta2021mobilevit}, making real-time deployment challenging.

To address these computational constraints, researchers have explored efficient transformer variants such as local/window-based attention \cite{Shi2022SSformerAL}, axial attention \cite{ho2019axial}, and lightweight attention mechanisms \cite{mehta2022separable, li2023rethinking, li2022efficientformer, pan2022edgevits, zhang2022topformer}. These methods partially alleviate complexity but still suffer from degradation in segmentation accuracy due to constrained receptive fields and reduced feature expressiveness, especially on high-resolution images (as demonstrated in~\cref{fig:comparison_of_all_models}).  To mitigate the computational costs associated with high-resolution processing in a transformer based approach, TopFormer \cite{zhang2022topformer} applies self-attention at a 1/64 downscaled resolution, reducing computational overhead at the cost of losing fine-grained spatial details. On the other hand, mobile-friendly CNN-based models ~\cite{ding2019acnet, ding2021repvgg, ma2018shufflenet, zhang2018shufflenet, howard2019searching} leverage depth-wise separable convolutions ~\cite{sandler2018mobilenetv2, howard2017mobilenets} for linear complexity scaling, but their local feature aggregation limits long-range contextual reasoning, ultimately reducing segmentation quality.

To bridge this gap, hybrid CNN-Transformer architectures have been proposed. Some methods integrate CNN backbones with Transformer decoders \cite{jin2021trseg, wang2022novel}, while others employ Transformer bottlenecks for global information processing \cite{graham2021levit, chen2022mobile, mehta2021mobilevit, zhang2022topformer}. 
While these approaches improve efficiency and expressiveness, they often struggle with an inherent trade-off: Transformer-based bottlenecks can introduce redundancy in global attention computation, while CNN-based bottlenecks, limited by local receptive fields, may fail to capture long-range dependencies effectively thus limiting their ability to simultaneously optimize efficiency, accuracy, and scalability in real-world segmentation tasks.

Building on these insights, we propose ContextFormer, a bottleneck-optimized segmentation framework that efficiently integrates global and local feature processing within a lightweight computational design. It comprises three key components: the Token Pyramid Extraction Module (TPEM), Transformer and Branched Depthwise Conv (Trans-BDC) block, and Feature Merging Module (FMM). TPEM constructs multi-scale token representations by fusing raw image intensity, gradient magnitude, and edge features via cascaded MobileNetV2 blocks~\cite{sandler2018mobilenetv2}, ensuring efficient feature encoding. The Trans-BDC block enhances contextual reasoning while preserving computational efficiency by integrating modulated branched depthwise convolutions with lightweight global attention. FMM adaptively redistributes spatial and channel-wise features across resolutions, refining hierarchical representations for robust segmentation. By synergizing these components within the bottleneck design, ContextFormer achieves superior efficiency-accuracy trade-offs, addressing key limitations in hybrid CNN-Transformer architectures.

To rigorously evaluate the efficacy of our proposed approach, we conduct a series of experiments on complex segmentation benchmarks, including ADE20K \cite{zhou2017scene}, Pascal Context \cite{mottaghi2014role}, CityScapes \cite{cordts2016cityscapes}, and COCO-Stuff \cite{caesar2018coco}. To further highlight the generalizability of our model, we validate its performance by conducting object detection experiments on the COCO \cite{lin2014microsoft} dataset.
The primary contributions of the work can be summarized as follows:
\begin{itemize}
  \item We propose ContextFormer, an innovative framework designed for efficient real-time semantic segmentation.
  \item We introduce a novel bottleneck design utilizing Trans-BDC blocks, which dynamically capture global dependencies and scale-aware semantics. These blocks, coupled with the TPEM for efficient multi-scale feature extraction and the FMM for adaptive feature integration, ensure robust hierarchical representations essential for dense prediction tasks.
  \item Our approach achieves state-of-the-art performance on the benchmark   datasets, significantly surpassing both mobile-optimized CNN and Transformer-based segmentation models by substantial margins on mIoU.
\end{itemize}

%% file: sec/2_Related_Work.tex
\section{Related Work} \label{sec:related_work}

\subsection{Efficient Convolutional Neural Networks}
Convolutional Neural Networks (CNNs) have dominated computer vision tasks over the last decade, mainly due to their translation equivalence and inductive bias. Recently, researchers have focused on making CNNs more efficient. MobileNet \cite{howard2019searching,howard2017mobilenets,sandler2018mobilenetv2}, for example, introduced the inverted bottleneck with depth-wise and point-wise convolutions, while ShuffleNet~\cite{ma2018shufflenet,zhang2018shufflenet} and IGCNet~\cite{zhang2017interleaved} used channel permutation for cross-group information flow. Further efficiency was achieved through models like ENet~\cite{paszke2016enet}, which reduces resolution early, and GhostNet~\cite{han2020ghostnet}, which optimizes depth-wise convolutions. Models like ERFNet~\cite{romera2017erfnet} and MobileNeXt~\cite{zhou2020rethinking} enhance speed and feature extraction, while TinyNet~\cite{dong2022tinynet}  and EfficientNet~\cite{li2019dfanet,li2019partial} explore scaling parameters to improve performance with less computation. For segmentation, high computational requirements often exceed device capabilities. To address this, DFANet \cite{li2019dfanet} and BiSeNet~\cite{yu2018bisenet} reduce cost through cross-level aggregation and dual-path architectures, while ICNet~\cite{zhao2018icnet} and ESPNet~\cite{mehta2019espnetv2} use multi-scale input and dilated convolutions. 
Unlike these approaches, which primarily emphasize local context, our method integrates multi-scale token processing and adaptive feature redistribution, enabling efficient capture of both local and global contextual information.

\subsection{Lightweight Vision Transformers}
Vision Transformers (ViTs)~\cite{dosovitskiy2020image} introduced transformer-based architectures to vision, surpassing CNNs in many image recognition tasks. Efforts to stabilize and enhance ViTs involved adding spatial inductive biases~\cite{dai2021coatnet,guo2022cmt} and adapting them for broader vision tasks~\cite{esser2021taming,zhang2022topformer} with more efficient self-attention mechanisms~\cite{dong2022cswin,zhu2023biformer}. To meet transformers' data demands, DeiT~\cite{touvron2021training} utilized token-based distillation, while T2T-ViT~\cite{yuan2021tokens} reduced token length through token aggregation. Swin Transformer~\cite{liu2021swin} achieved linear complexity by applying self-attention within local windows for greater efficiency. Despite their strong performance, ViTs often have high computational and memory costs, limiting their use on resource-constrained devices~\cite{mehta2021mobilevit,pan2022edgevits}. This inspired research into lightweight ViTs, like MobileFormer~\cite{chen2022mobile} and MobileViT~\cite{mehta2021mobilevit}, combining transformer-based self-attention with efficient MobileNet structures. Hybrid models, such as LeViT~\cite{graham2021levit} and EfficientFormer~\cite{li2022efficientformer}, blend CNNs and transformers for enhanced performance-speed balance. RTFormer~\cite{wang2022rtformer} uses a two-branch architecture based on cross- and GPU-friendly attention. TopFormer~\cite{zhang2022topformer} incorporates convolutional layers with transformer-based self-attention. These models typically employ a transformer encoder for global context through sequence processing, followed by a transformer or CNN decoder to integrate learned information. However, they often overlook the speed-accuracy tradeoff, leading to high computational demands. In contrast to standard ViT models, the proposed architecture efficiently leverages both CNNs and ViTs by pooling multi-scale tokens as inputs, achieving robust feature learning and improved generalizability.

%% file: sec/3_Proposed_Method.tex
\section{Proposed Method}
\label{sec:proposed_method}

\begin{figure*}
  \centering
  \includegraphics[width=\linewidth]{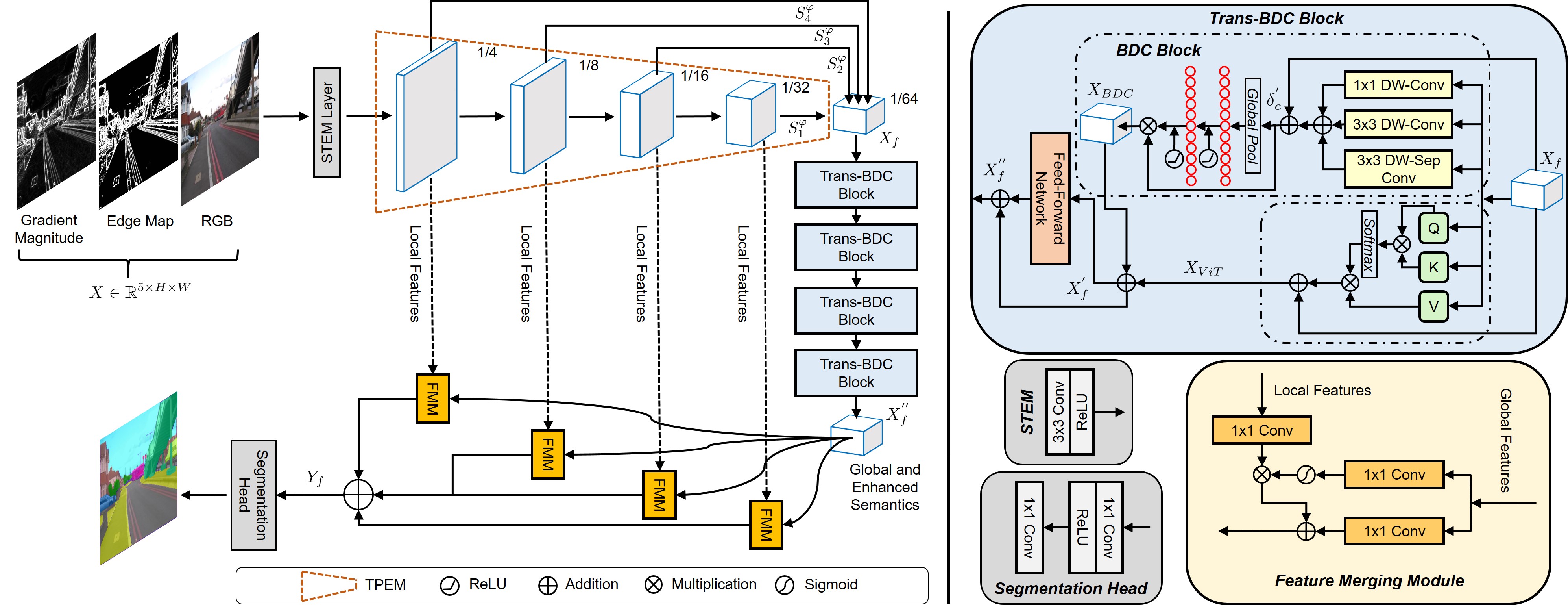}
  \caption{\textit{Holistic architecture of the proposed ContextFormer.} On (\textit{left}), the whole architecture is shown, whereas, on \textit{(right)} details of each of the blocks are depicted.
  }
  \label{fig:proposed_method}
\end{figure*}

This section presents ContextFormer, an efficient, robust, and powerful segmentation framework that avoids computationally intensive components. As shown in~\cref{fig:proposed_method}, it is composed of three key modules: (1) a Token Pyramid Extraction Module that captures high-resolution coarse features and low-resolution fine features and generates a feature pyramid, (2) a hybrid Trans-BDC block that efficiently generates scale-aware semantics, and (3) a lightweight Feature Merging Module that fuses these multi-level local and global semantic features to produce semantic segmentation mask via shifted gated mechanism. Next, the details of the model parts are explained below.%
\footnote{For better readability and ease, non-linear activation functions, batch normalization and dropout layers are not mentioned in the equations and figures.
}

\subsection{Token Pyramid Extraction Module}

Considering the efficiency constraints, the proposed Token Pyramid Extraction Module (TPEM) incorporates a series of stacked MobileNet blocks~\cite{sandler2018mobilenetv2}, specifically inverted residual blocks. 
However, unlike MobileNets, the proposed TPEM employs fewer blocks to establish a lightweight hierarchical feature representation. As demonstrated in~\cref{fig:proposed_method}, an input image ${X \in \mathbb{R}^{5 \times H \times W}}$ consists of stacked 3 channel (RGB) image along with the gradient magnitude and edge maps to enhance  the semantic richness, where $H$ and $W$ are the height and width of the image, respectively. The TPEM module initiates by passing this concatenated feature map via a series of MobileNetV2 blocks resulting in multi-scale features ${\{ S_1, S_2, ..., S_L\}}$ at different levels, where the number of levels is denoted by $L$ and the hierarchical feature at each level $L$ denoted as $
\frac{H}{2^{i+1}} \times \frac{W}{2^{i+1}} \times C_i$, where $i \in \{1, 2, 3, 4\}$. 

These multi-scale features encompass high-resolution coarse representations alongside low-resolution fine-grained details, which collectively enhance the efficacy of semantic segmentation.
To further minimize the computational overhead, an average pooling operation is applied to downsample the extracted tokens to a significantly reduced size, for instance, to $\frac{1}{64 \times 64}$ of the original input size. Finally, the same sized tokens from different scales are concatenated along the channel dimension to enhance the inter-channel dependencies and produce refined tokens for subsequent blocks. The overall operation of the TPEM is defined as:

\begin{equation}
X_{f} = \langle S_{1}^{\varphi}, S_{2}^{\varphi}, S_{3}^{\varphi}, S_{4}^{\varphi} \rangle
  \label{eq:3}
\end{equation}
Where, $\varphi$ denotes the average pooling operation applied to the corresponding feature map, and $\langle \cdot \rangle$ represents the concatenation operation. 
This combined local feature set is then passed to the Trans-BDC block to extract global and enhanced multi-scale semantics. 

\subsection{Trans-BDC Block}
The Trans-BDC block serves as an efficient bottleneck, integrating branched depthwise convolutions with lightweight self-attention to balance local and global feature extraction while reducing computational cost. As illustrated in ~\cref{fig:proposed_method}, to enhance efficiency, multi-scale features from the Token Pyramid Extraction Module (TPEM) undergo average pooling, ensuring a unified resolution before being processed in two parallel branches: a Branched Depthwise Convolution Module (BDC) for local feature extraction and a lightweight Vision Transformer (ViT) for global self-attention. Pooling at this stage reduces spatial redundancy, lowering FLOPs before deeper processing.

Basically, the BDC module employs three parallel depthwise convolutions: a $3 \times 3$ depthwise convolution for spatial semantics, a $1 \times 1$ depthwise convolution for cross-channel interactions, and a $3 \times 3$ depthwise separable convolution for efficient feature fusion. These outputs are aggregated and refined via a channel attention mechanism, enhancing contextual understanding with minimal computational overhead and the overall operation of BDC is defined as:
\begin{equation}
\delta_{c}^{'} = \xi_{3\times3}^{dw}(X_{f}) + \xi_{1\times1}^{dw}(X_{f}) + \xi_{1\times1}^{pw}(\xi_{3\times3}^{dw}(X_{f})) + X_{f}
%  \label{eq:5}
\end{equation}
\begin{equation}
X_{BDC} = \Gamma_{2}(\Gamma_{1}(\varphi_{g}(\delta_{c}^{'}))) \odot \delta_{c}^{'}
\label{eq:6}
\end{equation}

Where, $\varphi_{g}$ is the global average pooling layer, $\Gamma_{1}$ and $\Gamma_{2}$ are two fully-connected layers, $\odot$ represents the Hadamard product, and $\xi_{3\times3}^{dw}$, $\xi_{1\times1}^{dw}$, and $\xi_{1\times1}^{pw}$ denote $3 \times 3$ depth-wise, $1 \times 1$ depth-wise and $1 \times 1$ point-wise convolution operations, respectively.

Simultaneously, the ViT branch applies self-attention on the pooled features, but with key optimizations: $1 \times 1$ convolutions replace MLPs, reducing reshaping operations; lower-dimensional projections ($Q, K, V$ set to 16, 16, and 32, respectively) decrease memory footprint; batch normalization substitutes layer normalization for faster inference; and GELU activation is replaced with ReLU6 ~\cite{howard2017mobilenets}, further improving efficiency.
\begin{equation}
X_{ViT} = Attention(X_{f}) + X_{f}
\label{eq:self_attention}
\end{equation}

Where, $Attention =\text{softmax}\left(\frac{QK^T}{\sqrt{d_k}}\right)V$.
The outputs from the BDC and ViT branches are fused to form $X_{f}^{'}$, forming a refined feature representation that is then processed by a Feed-Forward Network (FFN) enhanced with depthwise convolutions and the overall operation of this block can be summarised as:
\begin{equation}
X_{f}^{''} = FFN(X_{f}^{'}) + X_{f}^{'}
\end{equation}

This reduces computational complexity while maintaining strong representational power. By leveraging branched feature extraction, token pooling, and optimized self-attention, Trans-BDC significantly lowers the computational burden in the bottleneck compared to conventional CNN or transformer-based architectures while preserving segmentation accuracy.

\subsection{Feature Merging Module and Segmentation Head}
After obtaining robust and enhanced semantics from the Trans-BDC block, Feature Merging Module (FMM) is incorporated to merge the complementary features from the Token Pyramid Extraction Module (TPEM) and Trans-BDC block. FMM module tries to alleviate the semantic gap of the different incoming information and allows the model to have both the local and global context information via its gated shifted mechanism. To this end, in FMM, firstly the  local features from TPEM are passed through a $1 \times 1$ convolution layer. 
Further, to get a weight matrix for the global features, a $1 \times 1$ convolution layer followed by Sigmoid activation function is applied on the global features/semantics from Trans-BDC block. Subsequently, the global semantics are integrated into the local tokens via the Hadamard product, and these global semantics are further summed, as shown in \cref{fig:proposed_method}, with the local features via a $1 \times 1$ convolution layer to focus on the relevant semantics. The overall operation of FMM module is summarized as follows: %Sizes of all three outputs are same.
\begin{equation}
Y_{f}= (\xi_{1\times1}^{c1}(X_{f})) \odot (\sigma(\xi_{1\times1}^{c2}(X_{f}^{''})))  + \xi_{1\times1}^{c3}(X_{f}^{''})
  \label{eq:9}
\end{equation}
Where, the first term before the  Hadamard product denotes the processing on the local tokens incoming from the {TPEM} module, the second term denotes the gated operation being performed on the features from the Trans-BDC block to obtain a weight matrix,  and $\sigma$ denotes Sigmoid activation function.

After obtaining enhanced features from the FMM, low-resolution features are upsampled to match the high-resolution features and summed element-wise. The resulting features are then passed through the Segmentation Head, comprising of two $1 \times 1$ convolution layers, to generate the final segmentation map.

%% file: sec/4_Experiments.tex
\section{Experiments}
\label{sec:experiments}

\subsection{Datasets and Measures}
We conduct the experiments on four benchmark datasets: ADE20K~\cite{zhou2017scene}, PASCAL Context~\cite{mottaghi2014role}, CityScapes~\cite{cordts2016cityscapes} and COCO-Stuff~\cite{caesar2018coco}.
In total, \textbf{ADE20K}~\cite{zhou2017scene} dataset consists of 25000 images, containing 150 class categories. The division of dataset is as follows: 20000 images for training, 2000 images for validation and 3000 images for testing.
The \textbf{PASCAL Context}~\cite{mottaghi2014role} dataset contains 1 background, and 59 semantic labels. Out of 10103 images in total, training and testing scene images are split into 4998 and 5105, respectively.
The \textbf{CityScapes}~\cite{cordts2016cityscapes} dataset consists of 19 fine class annotations. 2975 images are taken into account for training and 500 images for validation/testing.
Pixel-level stuff annotations were applied on COCO dataset for augmentation, resulting in \textbf{COCO-Stuff}~\cite{caesar2018coco} dataset. From COCO dataset, 10000 images are picked, where 9000 images are considered for training and 1000 images for testing.

Following the recent literature~\cite{kang2024metaseg,cavagnero2024pem,zhang2022topformer} we report the results using the standard common measures: Mean Intersection over Union (mIoU) for segmentation accuracy, Giga Floating Point Operations per Second (GFLOPs), latency and number of parameters. Details for \textit{object detection} task are provided in \cref{subsec:object_detection}.

\subsection{Implementation Details}
Our implementation is based on PyTorch and MMSegmentation toolbox~\cite{mmseg2020}. All the models are first pretrained on ImageNet-1K dataset~\cite{deng2009imagenet}, including our proposed ContextFormer\footnote{On ImageNet-1K dataset, ContextFormer achieves 66.4\% Top-1 accuracy. Details are provided in the supplementary material.
} model. Further, they are fined-tuned on semantic segmentation datasets. Batch-Normalization layers are used after almost each convolution layer except the last output layer.  
For the \textbf{ADE20K} dataset, we follow data augmentations same as in \cite{xie2021segformer} for fair comparison. Moreover, for ADE20K dataset, we use batch size 16, and 160K scheduler by following \cite{xie2021segformer} and \cite{zhang2022topformer}. Various augmentations have been applied such as random scaling, random cropping, random horizontal flip, random resize, etc. For the \textbf{CityScapes} dataset, the same data augmentations are followed as in \cite{xie2021segformer,zhang2022topformer}. Images are resized and rescaled with same crop size \ie $1024 \times 512$. It should be noted that for all datasets and models, we set $1.2 \times 10^{-4}$ as the initial learning rate with  the weight decay set to $0.01$. However, the initial learning rate for the CityScapes dataset is set to $3 \times 10^{-4}$. 
For the \textbf{PASCAL Context} and \textbf{COCO-Stuff} datasets, 80K training iterations are implemented. Additionally, for both datasets, the same data augmentations and training settings are incorporated as in \cite{mmseg2020} and the training images are resized and cropped to $512 \times 512$ and $480 \times 480$ for COCO-Stuff and PASCAL Context datasets, respectively. 
\begin{table*}
  \caption{Quantitative results of various models on ADE20K \textit{validation} set~\cite{zhou2017scene}. * indicates results from models trained with $448 \times 448$ input size. For fair comparison with SegFormer, TopFormer and SeaFormer batch size of 16 is considered for ContextFormer. Whereas, batch size of 32 is considered for CNN based models. GFLOPs are reported for input resolution of $512 \times 512$. `-' means the unreported results. Single-scale inference is used to report mIoU.
  }
  \label{tab:results_ade20k}
  \centering
  {
  \setlength{\tabcolsep}{8pt}
  \begin{tabular}{@{}l|l|c|c|c|c@{}}
    \toprule
    Models&Encoder&mIoU&GFLOPs&Parameters&Latency(ms)\\
    \midrule
    PSPNet \cite{zhao2017pyramid}&MobileNetV2~\cite{sandler2018mobilenetv2}&29.6&52.2&13.7M&426\\
    FCN-8s \cite{long2015fully}&MobileNetV2~\cite{sandler2018mobilenetv2}&19.7&39.6&9.8M&406\\
    Semantic FPN \cite{kirillov2019panoptic}&ConvMLP-S~\cite{li2023convmlp}&35.8&33.8&12.8M&311\\
    DeepLabV3+ \cite{chen2018encoder}&EfficientNet~\cite{tan2019efficientnet}&36.2&26.9&17.1M&388\\
    DeepLabV3+ \cite{chen2018encoder}&MobileNetV2~\cite{sandler2018mobilenetv2}&38.1&25.8&15.4M&414\\
    Lite-ASPP \cite{chen2018encoder}&ResNet18~\cite{he2016deep}&37.5&19.2&12.5M&259\\
    DeepLabV3+ \cite{chen2018encoder}&ShuffleNetv2-1.5x~\cite{ma2018shufflenet}&37.6&15.3&16.9M&384\\
    HRNet-Small \cite{yuan2020object}&HRNet-W18-Small~\cite{yuan2020object}&33.4&10.2&4.0M&256\\
    SegFormer \cite{xie2021segformer}&MiT-B0~\cite{xie2021segformer}&37.4&8.4&3.8M&308\\
    Lite-ASPP \cite{chen2018encoder}&MobileNetV2~\cite{sandler2018mobilenetv2}&36.6&4.4&2.9M&94\\
    R-ASPP \cite{sandler2018mobilenetv2}&MobileNetV2~\cite{sandler2018mobilenetv2}&32.0&2.8&2.2M&71\\
    HR-NAS-B \cite{ding2021hr}&Searched~\cite{ding2021hr}&34.9&2.2&3.9M&-\\
    LR-ASPP \cite{howard2019searching}&MobileNetV3-Large~\cite{howard2019searching}&33.1&2.0&3.2M&51\\
    HR-NAS-A \cite{ding2021hr}&Searched~\cite{ding2021hr}&33.2&1.4&2.5M&-\\
    LR-ASPP \cite{howard2019searching}&MobileNetV3-Large-reduce~\cite{howard2019searching}&32.3&1.3&1.6M&33\\
    SeaFormer \cite{wan2023seaformer}&SeaFormer-T~\cite{wan2023seaformer}&34.7&0.6&1.7M&15\\
    TopFormer* \cite{zhang2022topformer}&TopFormer-T~\cite{zhang2022topformer}&32.5&0.5&1.4M&14\\
    TopFormer \cite{zhang2022topformer}&TopFormer-T~\cite{zhang2022topformer}&32.8&0.6&1.4M&15\\
    \rowcolor{gray!40}
    ContextFormer* (Ours)&TPEM&34.2&0.5&1.7M&15\\
    \rowcolor{gray!40}
    ContextFormer (Ours)&TPEM&34.8&0.6&1.7M&15\\
    \rowcolor{gray!40}
   ContextFormer*$_{(GME)}$ (Ours)&TPEM&34.7&0.5&1.7M&15\\
    \rowcolor{gray!40}
    ContextFormer$_{(GME)}$ (Ours)&TPEM&35.0&0.6&1.7M&15\\   
  \bottomrule
  \end{tabular}
  }
\end{table*}

\subsection{Experiments on ADE20K}
\begin{table}
  \caption{Comparison of ContextFormer with more \textit{recent efficient models} on ADE20K dataset~\cite{zhou2017scene}. * indicates results from models trained with $448 \times 448$ input size. 
  }
    \label{tab:results_sota_ade20k}
    \centering
    \resizebox{\columnwidth}{!}{%
    \begin{tabular}{@{}l|l|c|c|c@{}}
    \toprule
    Models&Encoder&mIoU&GFLOPs&Parameters\\
    \midrule
    PEM~\cite{cavagnero2024pem}&STDC1~\cite{cavagnero2024pem}&39.6&16.0&17.0M\\
    FeedFormer-B0 \cite{shim2023feedformer}&MiT-B0~\cite{xie2021segformer}&39.2&7.8&4.5M\\
    SegNeXt \cite{guo2022segnext}&SegNeXt-T~\cite{guo2022segnext}&41.1&6.6&4.3M\\
    U-MixFormer \cite{yeom2023u}&MiT-B0~\cite{xie2021segformer}&41.2&6.1&6.1M\\
    MegaSeg \cite{kang2024metaseg}&MegaSeg-T~\cite{kang2024metaseg}&42.4&5.5&4.7M\\
    CGRSeg \cite{ni2024context}&CGRSeg-T~\cite{ni2024context}&43.6&4.0&9.4M\\
    SeaFormer \cite{wan2023seaformer}&SeaFormer-T~\cite{wan2023seaformer}&34.7&0.6&1.7M\\
    TopFormer* \cite{zhang2022topformer}&TopFormer-T~\cite{zhang2022topformer}&32.5&0.5&1.4M\\
    TopFormer \cite{zhang2022topformer}&TopFormer-T~\cite{zhang2022topformer}&32.8&0.6&1.4M\\
    \rowcolor{gray!40}
    ContextFormer* (Ours)&TPEM&34.2&0.5&1.7M\\
    \rowcolor{gray!40}
    ContextFormer (Ours)&TPEM&34.8&0.6&1.7M\\
    \rowcolor{gray!40}
    ContextFormer*$_{(GME)}$ (Ours)&TPEM&34.7&0.5&1.7M\\
    \rowcolor{gray!40}
    ContextFormer$_{(GME)}$ (Ours)&TPEM&35.0&0.6&1.7M\\
  \bottomrule
  \end{tabular}
  }
\end{table}

As illustrated in~\cref{tab:results_ade20k}, the proposed model is compared with state-of-the-art approaches, consisting of efficient CNNs \cite{he2016deep,howard2019searching,sandler2018mobilenetv2,sun2019deep} and lightweight ViTs \cite{ding2021hr,kirillov2019panoptic,ma2018shufflenet,xie2021segformer,zhang2022topformer}, on the validation set of the ADE20K dataset. For performance and efficiency evaluation, mIoU, GFLOPs, parameters, and latency are taken into account. The proposed model is trained for the following input resolutions: $512 \times 512$ and $448 \times 448$ on the ADE20K dataset.

DeepLabV3+ with MobileNetV2 as the encoder achieved the best mIoU \ie 38.1\%. Notwithstanding, it encompasses a substantially higher computational burden measured in GFLOPs and latency, contrasting with the proposed model, which achieves comparable efficacy with 25.2 fewer GFLOPs—representing a reduction of 97.67\%—and 13.7M (88.96\%) fewer parameters while having just 15ms latency, signifying a decrease in complexity.

Among the CNN-based baselines, the most popular model utilizing MobileNetV3-Large \cite{howard2019searching} as the encoder and LR-ASPP as the decoder strikes an effective balance between the computational complexity (2.0 GFLOPs, 51ms latency) and accuracy (33.1\% mIoU). To make this model more efficient, a reduced version, called MobileNetV3-Large-reduce, was used. Our proposed ContextFormer (35.0\% mIoU) surpasses both the LR-ASPP models in accuracy by 1.9\% and 2.7\%, respectively, while having much lower latency (15ms), thus proving its robustness. This represents a latency reduction of 70.59\% compared to LR-ASPP and 54.55\% compared to LR-ASPP-reduce.

In ViT-based models, HR-NAS-B \cite{ding2021hr} integrates Transformer blocks into the HRNet \cite{wang2020deep} architecture using neural architecture search, achieving an effective balance between computational complexity (2.2 GFLOPs) and accuracy (34.9\% mIoU). SegFormer has shown good performance of 37.4\% mIoU. Nevertheless, the proposed model achieves an mIoU of 35.0\% while having 92.9\% and 2.1M (56.4\%) fewer GFLOPs and parameters, respectively, compared to SegFormer, and 0.1\% higher mIoU than HR-NAS-B. Moreover, ContextFormer demonstrates a 72.7\% reduction in GFLOPs compared to HR-NAS-B and a 92.9\% reduction compared to SegFormer.
When compared to TopFormer \cite{zhang2022topformer}, which achieves 32.8\% mIoU with 0.6 GFLOPs and 1.4M parameters, the proposed model delivers a 2.2\% higher mIoU while maintaining the same GFLOPs and slightly higher parameters (1.7M). Similarly, when compared to SeaFormer \cite{wan2023seaformer}, which achieves 34.7\% mIoU with 0.6 GFLOPs and 1.7M parameters, the proposed model achieves a 0.3\% higher mIoU without increasing computational costs or parameters. These results highlight the superior performance of the proposed ContextFormer in balancing accuracy and efficiency.

For further demonstrating the effectiveness of the proposed ContextFormer, we also report the results on some recent SOTA semantic segmentation approaches in different settings. It can be clearly seen in \cref{tab:results_sota_ade20k} that unlike other approaches, our proposed model achieves a better trade-off between performance and efficiency.

\subsubsection{Visual Results}
\begin{figure}
  \centering
  \includegraphics[width=\linewidth]{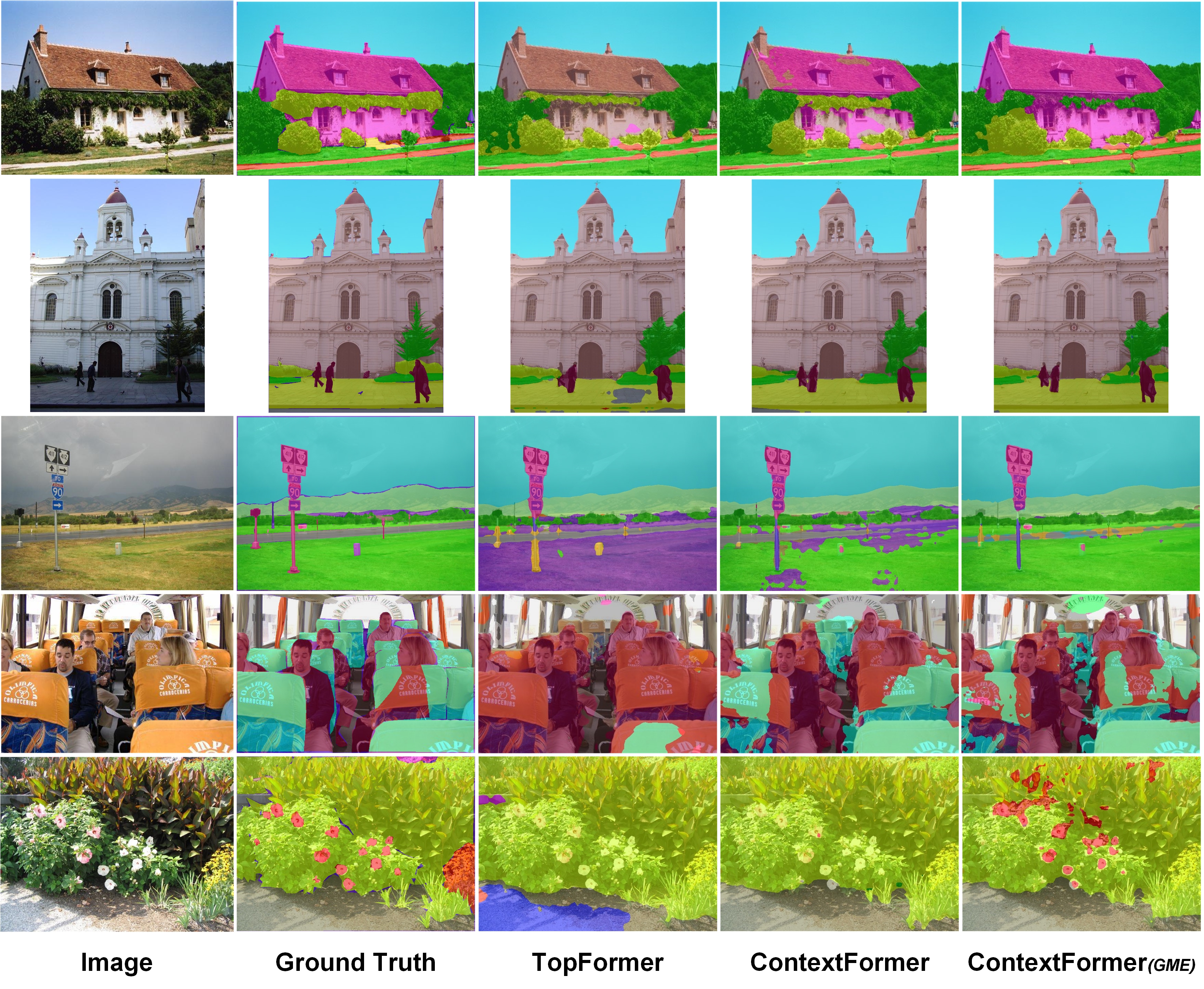}
  \caption{Visual results on the ADE20K validation set. The results highlight the proposed model's effectiveness in producing high-quality segmentation maps with improved spatial consistency.
  }
  \label{fig:proposed_method_visual_results}
\end{figure}

~\cref{fig:proposed_method_visual_results} illustrates the qualitative results of our proposed model, compared alongside the original images, ground truth annotations, and the segmentation results produced by TopFormer \cite{zhang2022topformer}. The input images used for this evaluation are sourced from the validation set of the ADE20K benchmark dataset, which is renowned for its challenging and diverse scenes. A detailed examination of the visual outputs reveals that our model consistently delivers more precise segmentation results, capturing finer details and producing more realistic predictions compared to TopFormer. These improvements are particularly evident in regions with intricate boundaries and small objects, where TopFormer often exhibits over-smoothing or segmentation errors. The superior performance of both of our models underscores its robustness and ability to effectively learn and generalize complex visual features from the dataset. Such results not only demonstrate the model's capacity for capturing subtle details but also highlight its potential for real-world applications requiring high-quality semantic segmentation with improved edge fidelity and accuracy.

\subsubsection{Ablation Study}
As elaborated in~\cref{tab:results_ade20K_ablation}, we analyze the contribution of the Trans-BDC block to the overall performance of the model, focusing on its individual components and their collective impact. The study is divided into two parts, examining the \textbf{B}ranched \textbf{D}epthwise\textbf{C}onv (BDC) block's architecture and the integration of Vision Transformer (ViT) components.

\textbf{Effects of BDC Block Design:}
The upper half of the study highlights the incremental benefits of specific design choices within the BDC block. Starting with the baseline configuration, which includes a $3\times3$ depth-wise convolution layer, we observe that augmenting it with an additional $1\times1$ depth-wise convolution layer enhances the model's segmentation accuracy by 0.2\%. This improvement, though marginal, demonstrates the effectiveness of fine-grained feature extraction at multiple scales. The inclusion of a third parallel branch in the BDC block, comprising a $3\times3$ depth-wise separable convolution, significantly boosts performance by 3.3\%. This substantial improvement underscores the importance of capturing spatial hierarchies and multi-scale contextual information for robust feature representation, thus capturing the global dependencies. Additionally, the integration of a channel attention layer further refines the extracted features, albeit with a smaller but meaningful performance gain, highlighting its utility in emphasizing critical feature channels.

\textbf{Effect of Trans-BDC Block and GME Map:} In the lower half of the study, we evaluate the synergy between the ViT backbone and the BDC block. Initially, the model demonstrates a substantial 32.7\% mIoU improvement when utilizing ViT alone, even without the BDC block, validating ViT's efficacy in enhancing semantic feature representation. However, the stepwise inclusion of depth-wise convolution layers from the BDC block markedly improves the model's performance. The addition of each BDC component further strengthens ViT's capability to capture local and global context, resulting in a cumulative performance improvement of approximately 2.1\%, as detailed. These results provide compelling evidence that the BDC block not only complements ViT but also enhances its effectiveness, resulting in a robust architecture that excels in overall segmentation tasks. 

Furthermore, the inclusion of edge maps alongside the RGB image further amplifies the model's performance by 0.2\%, thus effectively capturing sharp and fine-grained features within the image. This proves that the edge maps act as a complementary input, guiding the segmentation process by emphasizing object boundaries and delineating intricate structures. Moreover, our explicit approach can generalize to any edge detector (e.g. Canny Edge) in a zero-shot way. \cref{tab:contextformer_comparison_with_canny_edge} highlights our consistent performance, with Canny introducing additional latency without providing significant improvements over Sobel, our plain approach.

\begin{table}
  \caption{Ablation studies of ContextFormer on the ADE20K \textit{validation} set~\cite{zhou2017scene}.
  }
  \vspace{-2mm}
  \label{tab:results_ade20K_ablation}
  \centering
  \resizebox{\columnwidth}{!}{%
  \begin{tabular}{@{}c|c|c|c|c|c|c|c|c@{}}
    \toprule
    \multicolumn{6}{c|}{ContextFormer (Trans-BDC)} & mIoU & GFLOPs & Parameters\\
    \cline{1-6}
 ViT$_{block}$ & $3\times3_{dw}$& $1\times1_{dw}$& $3\times3_{dw.sep}$& C-Attn & GME & &\\
 \midrule
    & \checkmark &  &  &  &  & 25.4 & 0.55 & 1.02M\\
    & \checkmark & \checkmark &  &  &  & 25.6 & 0.55 & 1.10M \\
    & \checkmark & \checkmark & \checkmark &  &   & 28.7 & 0.56 & 1.29M \\
    & \checkmark & \checkmark & \checkmark & \checkmark &  & 29.5 & 0.56 & 1.38M\\
    & \checkmark & \checkmark & \checkmark & \checkmark & \checkmark & 29.8 & 0.56 & 1.38M\\
    \checkmark &  &  &  &  &  & 32.7 & 0.54 & 1.41M\\
    \checkmark & \checkmark &  &  &  &  & 32.8 &  0.57 & 1.42M  \\
    \checkmark &  \checkmark & \checkmark &  &  &  & 32.9 & 0.57 & 1.43M\\
    \checkmark &  \checkmark & \checkmark & \checkmark &  &  & 33.8 & 0.58 & 1.61M \\
    \checkmark &  \checkmark & \checkmark & \checkmark & \checkmark &  & 34.8 & 0.58 & 1.68M\\
    \rowcolor{gray!40}
    \checkmark &  \checkmark & \checkmark & \checkmark & \checkmark & \checkmark & 35.0 & 0.58 & 1.68M\\
 \bottomrule
    \end{tabular}
    }
\end{table}

\begin{table}[t]
\scriptsize
\footnotesize
  \caption{Comparison of our model including GME variants.}
  \vspace{-2mm}
  \label{tab:contextformer_comparison_with_canny_edge}
  \centering
  \begin{tabular}{@{}c|c|c|c|c@{}}
    \toprule
        GME (Variations) & mIoU & GFLOPs & Parameters & Latency(ms)  \\
    \midrule
        Canny & 31.6 & 0.6 & 1.7M & 18.8 \\
        Sobel+Canny & 35.0 & 0.6 & 1.7M & 19.2 \\
        \rowcolor{gray!40}
        Sobel (Ours) & 35.0 & 0.6 & 1.7M & 15.1  \\
    \bottomrule
  \end{tabular}
\end{table}

\subsection{Results on PASCAL Context}
~\cref{tab:results_pascal_context} highlights the comparative performance on the PASCAL Context dataset for 59-class and 60-class configurations. DeepLabV3+ with ENet-s16 and MobileNetV2-s16 backbones achieves mIoU$^{59}$/mIoU$^{60}$ scores of 43.07\%/39.19\% and 42.34\%/38.59\%, respectively, but with high computational costs of 23.00 and 22.24 GFLOPs. In contrast, the proposed method delivers competitive mIoU$^{59}$/mIoU$^{60}$ scores of 41.85\%/37.78\%, with 22.51 (97.87\%) fewer GFLOPs, showcasing its efficiency.

LR-ASPP achieves mIoU$^{59}$/mIoU$^{60}$ scores of 38.02\%/35.05\% using only 2.04 GFLOPs, but the proposed method surpasses it by 3.83\%/2.73\% with a marginal reduction of 1.55 (75.98\%) GFLOPs. Similarly, TopFormer balances performance and efficiency but is outperformed by the proposed method, which achieves 1.46\% and 1.37\% higher mIoU$^{59}$ and mIoU$^{60}$, respectively, while reducing GFLOPs by 0.04 (7.55\%).
When compared to SeaFormer, which achieves mIoU$^{59}$/mIoU$^{60}$ scores of 41.49\%/37.27\% with 0.51 GFLOPs, the proposed method achieves 0.36\% and 0.51\% higher mIoU$^{59}$ and mIoU$^{60}$, respectively, while reducing GFLOPs by 0.02 (3.92\%).
These results establish the proposed method as achieving the optimal trade-off between performance and computational efficiency.

\begin{table}
  \caption{Results on PASCAL Context \textit{test} set~\cite{mottaghi2014role}.
  }
  \vspace{-2mm}
  \label{tab:results_pascal_context}
  \centering
  \resizebox{\columnwidth}{!}{%
  \begin{tabular}{@{}l|l|c|c|c@{}}
    \toprule
    Methods & Backbone & mIoU$^{59}$ & mIoU$^{60}$ & GFLOPs \\
    \midrule
    DeepLabV3+~\cite{chen2018encoder} & ENet-s16~\cite{paszke2016enet} & 43.07 & 39.19 & 23.00 \\
    DeepLabV3+~\cite{chen2018encoder} & MobileNetV2-s16~\cite{sandler2018mobilenetv2} & 42.34 & 38.59 & 22.24 \\
    LR-ASPP~\cite{howard2019searching} & MobileNetV3-s16~\cite{howard2019searching} & 38.02 & 35.05 & 2.04 \\
    TopFormer~\cite{zhang2022topformer} & TopFormer-T~\cite{zhang2022topformer} & 40.39 & 36.41 & 0.53 \\
    SeaFormer \cite{wan2023seaformer} & SeaFormer-T~\cite{wan2023seaformer} & 41.49 & 37.27 & 0.51 \\
    \rowcolor{gray!40}
    ContextFormer (Ours) & TPEM & 41.84 & 37.49 & 0.47 \\
    \rowcolor{gray!40}
    ContextFormer$_{(GME)}$ (Ours) & TPEM & 41.85 & 37.78 & 0.49 \\
      \bottomrule
    \end{tabular}
    }
\end{table}

\subsection{Results on CityScapes}
To further validate the efficacy of the proposed method, we summarize the results on the CityScapes dataset in~\cref{tab:results_cityscapes}. While PSPNet and FCN, both with MobileNetV2 encoders, achieve mIoU scores of 70.2 and 61.5, they require 423.4 and 317.1 GFLOPs, respectively. In contrast, the proposed method delivers 6.7\% higher mIoU than FCN with a 315.9 (99.62\%) GFLOPs reduction and performs comparably to PSPNet with 422.2 (99.71\%) fewer GFLOPs.

L-ASPP achieves the highest mIoU of 72.7 among CNN-based models, but at the cost of 11.4 (90.48\%) additional GFLOPs compared to our method. Similarly, SegFormer, a ViT-based model, achieves a comparable mIoU of 71.9 but requires 16.5 (93.22\%) more GFLOPs. The proposed method also surpasses TopFormer with 1.7\% higher mIoU at the same computational cost.
Additionally, when compared to SeaFormer, which achieves 66.8 mIoU with 1.2 GFLOPs, the proposed method delivers 1.4\% higher mIoU without additional computational cost. These results demonstrate the proposed method's superior balance of performance and efficiency.

\begin{table}
  \caption{Results on CityScapes \textit{validation} set~\cite{cordts2016cityscapes}.
  }
  \vspace{-2mm}
  \label{tab:results_cityscapes}
  \centering
  \resizebox{\columnwidth}{!}{%
  \begin{tabular}{@{}l|l|c|c@{}}
    \toprule
    Methods & Encoder & mIoU & GFLOPs \\
    \midrule
    PSPNet~\cite{zhao2017pyramid} & MobileNetV2~\cite{sandler2018mobilenetv2} & 70.2 & 423.4 \\
    FCN~\cite{long2015fully} & MobileNetV2~\cite{sandler2018mobilenetv2} & 61.5 & 317.1 \\
    SegFormer~\cite{xie2021segformer} & MiT-B0~\cite{xie2021segformer} & 71.9 & 17.7 \\
    L-ASPP~\cite{chen2018encoder} & MobileNetV2~\cite{sandler2018mobilenetv2} & 72.7 & 12.6 \\
    LR-ASPP~\cite{howard2019searching} & MobileNetV3-Large~\cite{howard2019searching} & 72.4 & 9.7 \\
    LR-ASPP~\cite{howard2019searching} & MobileNetV3-Small~\cite{howard2019searching} & 68.4 & 2.9 \\
    TopFormer~\cite{zhang2022topformer} & TopFormer-T~\cite{zhang2022topformer} & 66.5 & 1.2 \\
    SeaFormer \cite{wan2023seaformer}&SeaFormer-T~\cite{wan2023seaformer} & 66.8 & 1.2 \\
    \rowcolor{gray!40}
    ContextFormer (Ours) & TPEM & 68.0 & 1.2 \\
    \rowcolor{gray!40}
    ContextFormer$_{(GME)}$ (Ours) & TPEM & 68.2 & 1.2 \\
      \bottomrule
    \end{tabular}
    }
\end{table}

\subsection{Results on COCO-Stuff}
~\cref{tab:results_coco_stuff} summarizes the performance of various models on the COCO-Stuff dataset. While DeepLabV3+ with EfficientNet-s16 achieves the highest mIoU of 31.45, it incurs substantial computational costs with 27.10 GFLOPs. In contrast, the proposed method achieves comparable accuracy with 26.50 (97.8\%) fewer GFLOPs, demonstrating its efficiency. Similarly, compared to DeepLabV3+ with MobileNetV2-s16, which consumes 25.90 GFLOPs for an mIoU of 29.88, the proposed method achieves similar performance with 25.30 (97.7\%) fewer GFLOPs, highlighting its scalability.

Additionally, the proposed method matches PSPNet’s accuracy while significantly reducing computational demand and surpasses TopFormer with a 0.92\% higher mIoU and 0.04 (6.25\%) fewer GFLOPs. When compared to SeaFormer, the proposed method achieves 0.02\% higher mIoU with 0.02 (3.23\%) fewer GFLOPs. These results establish the proposed method as a robust and efficient solution, achieving an optimal balance between performance and computational cost for real-world applications.

\begin{table}
  \caption{Results on COCO-Stuff \textit{test} set~\cite{caesar2018coco}.
  }
  \vspace{-2mm}
  \label{tab:results_coco_stuff}
  \centering
  \resizebox{\columnwidth}{!}{%
  \begin{tabular}{@{}l|l|c|c@{}}
    \toprule
    Methods & Encoder & mIoU & GFLOPs \\
    \midrule
    PSPNet~\cite{zhao2017pyramid} & MobileNetV2-s8~\cite{sandler2018mobilenetv2} & 30.14 & 52.94 \\
    DeepLabV3+~\cite{chen2018encoder} & EfficientNet-s16~\cite{tan2019efficientnet} & 31.45 & 27.10 \\
    DeepLabV3+~\cite{chen2018encoder} & MobileNetV2-s16~\cite{sandler2018mobilenetv2} & 29.88 & 25.90 \\
    LR-ASPP~\cite{howard2019searching} & MobileNetV3-s16~\cite{howard2019searching} & 25.16 & 2.37 \\
    TopFormer~\cite{zhang2022topformer} & TopFormer-T~\cite{zhang2022topformer} & 28.34 & 0.64 \\
    SeaFormer \cite{wan2023seaformer} & SeaFormer-T~\cite{wan2023seaformer} & 29.24 & 0.62  \\
    \rowcolor{gray!40}
    ContextFormer (Ours) & TPEM & 29.23 & 0.58 \\
    \rowcolor{gray!40}
    ContextFormer$_{(GME)}$ (Ours) & TPEM & 29.26 & 0.60 \\
      \bottomrule
    \end{tabular}
    }
\end{table}

\subsection{Object Detection}
\label{subsec:object_detection}
To further validate the generalization capability of our proposed method, we conducted object detection experiments using the COCO \cite{lin2014microsoft} dataset to evaluate its performance in a downstream task. The COCO dataset comprises 118K training images, 5K validation images, and 20K test images. For the detection framework, we employed RetinaNet \cite{lin2017focal}, a one-stage model, and integrated ContextFormer$_{(GME)}$ as the backbone to generate a feature pyramid across multiple scales. All models were trained on the train2017 split for 12 epochs using ImageNet pretrained weights and were evaluated on the val2017 set. As evident from \cref{tab:contextformer_object_detection_results}, our model excels in the detection task, showcasing the robust generalization capability of our method.

\begin{table}[t]
\footnotesize
  \caption{Object detection results on COCO dataset.}
  \vspace{-2mm}
  \label{tab:contextformer_object_detection_results}
  \centering
  \resizebox{\columnwidth}{!}{%
  \begin{tabular}{@{}l|c|c|c@{}}
    \toprule
    Backbone & mAP & GFLOPs & Parameters  \\
    \midrule
    ShuffleNetV2 \cite{ma2018shufflenet} & 25.9 & 161 & 10.4M \\
    MobileNetV3 \cite{howard2019searching} & 27.2 & 162 & 12.3M \\
    TopFormer-T \cite{zhang2022topformer} & 27.1 & 160 & 10.5M \\
    SeaFormer-T \cite{wan2023seaformer}& 31.5 & 160 & 10.9M \\
    \rowcolor{gray!40}
    ContextFormer$_{(GME)}$ (Ours) & 31.6 & 160 & 10.9M  \\
     \bottomrule
  \end{tabular}
  }
\end{table}

%% file: sec/5_Conclusion.tex
\section{Conclusion}
ContextFormer is a hybrid framework that combines the strengths of CNNs and ViTs within the bottleneck to balance efficiency, accuracy, and robustness for real-time semantic segmentation. Its efficiency is driven by three synergistic modules: TPEM, the Trans-BDC block, and the FMM blocks. Extensive experiments across four benchmark datasets demonstrate that ContextFormer achieves a superior balance between accuracy and computational efficiency compared to existing methods. Its versatile design makes it adaptable to various applications, paving the way for future research into efficient ViT-CNN hybrid models across diverse domains.

%% file: sec/X_suppl.tex
\clearpage
\setcounter{page}{1}
\maketitlesupplementary

\setcounter{section}{0}
\setcounter{figure}{0}    
\setcounter{table}{0}

\renewcommand{\thetable}{\Alph{table}}
\renewcommand{\thefigure}{\Alph{figure}}
\renewcommand{\thesection}{\arabic{section}}

\renewcommand{\theHsection}{suppl.\thesection}

In this supplementary, we illustrate the ImageNet pre-training and results for the task of image classification (\cref{sec:imagenet_pretraining}), model architecture with details (\cref{sec:network_details}), design of Feed-Forward Network (\cref{sec:fnn_details}), visual results (\cref{sec:visual_results}), and limitations with future work (\cref{sec:limitations_futurework}).

\section{ImageNet Pre-training} 
\label{sec:imagenet_pretraining}

\begin{figure}
  \centering
  \includegraphics[width=\linewidth]{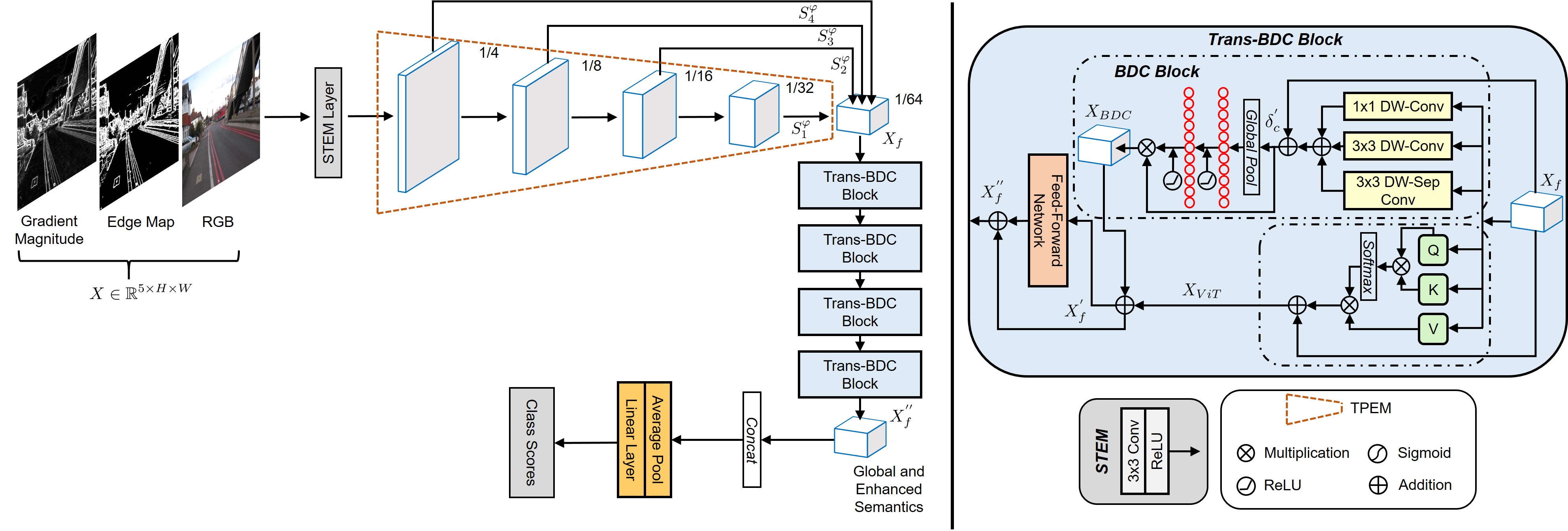}
  \caption{The architecture of the proposed ContextFormer model for the task of image classification.
  }
  \label{fig:proposed_method_classification}
\end{figure}

For a fair comparison, we initialize the ContextFormer model using pre-trained parameters from ImageNet.  As demonstrated in \cref{fig:proposed_method_classification}, the classification framework of ContextFormer integrates an average pooling layer followed by a linear layer, leveraging global semantic representations to generate class scores.  
Given the low resolution of the input images ($224 \times 224$), the target resolution of the input tokens for the Trans-BDC block is configured to be $\frac{1}{64} \times \frac{1}{64}$ of the input dimensions.
Quantitative results of the proposed ContextFormer model on the ImageNet-1K dataset are shown in~\cref{tab:results_imagenet1k}.
\begin{table}[h]
  \caption{ContextFormer results for ImageNet classification.
  }
  \label{tab:results_imagenet1k}
  \centering
  \resizebox{\columnwidth}{!}{%
  \begin{tabular}{@{}c|c|c|c|c@{}}
    \toprule
    Method & Input Size & Top-1 Accuracy(\%) & GFLOPs & Parameters \\
    \midrule
    ContextFormer & $224\times224$ & 66.1 & 0.13 & 1.79M \\
    ContextFormer$_{(GME)}$ & $224\times224$ & 66.4 & 0.13  & 1.79M  \\
      \bottomrule
    \end{tabular}
    }
\end{table}

\section{Detailed Network Structure} 
\label{sec:network_details}
The detailed network structure of the proposed ContextFormer model for the efficient real-time semantic segmentation task is provided in~\cref{tab:architectural_details_ContextFormer}. Where, $dw$ and $dw.sep$ denote depth-wise and depth-wise separable convolutions. Moreover, N, H and T denote the number of blocks, number of heads and number of target channels, respectively. The input image resolution taken into account is $512 \times 512$.

\begin{table}
  \caption{Architectural details of ContextFormer model. For input resolution of $512 \times 512$.
  }
  \label{tab:architectural_details_ContextFormer}
  \centering
  \resizebox{\columnwidth}{!}{%
  \begin{tabular}{@{}c|c|c|c|c|c|c@{}}
    \toprule
    Stage & \multicolumn{5}{c|}{Details} & Output Resolution \\
    \cline{2-6}
    \cline{2-5}
 & layer & kernel size & expand ratio & output channels & stride \\
 \midrule
 Stem & Conv & 3 & - & 16 & 2 & \\
   & MobileNetV2 & 3 & 1 & 16 & 1 & $256 \times 256$\\
 \midrule
 & MobileNetV2 & 3 & 4 & 16 & 2 \\
 & MobileNetV2 & 3 & 3 & 16 & 1 & $128 \times 128$ \\
 \cline{2-7}
 & MobileNetV2 & 5 & 3 & 32 & 2 \\
 TPEM & MobileNetV2 & 5 & 3 & 32 & 1 & $64 \times 64$\\
 \cline{2-7}
 & MobileNetV2 & 3 & 3 & 64 & 2 \\
 & MobileNetV2 & 3 & 3 & 64 & 1 & $32 \times 32$\\
 \cline{2-7}
 & MobileNetV2 & 5 & 6 & 96 & 2 \\
 & MobileNetV2 & 5 & 6 & 96 & 1 & $16 \times 16$\\
 \midrule
 \midrule
  & \multicolumn{5}{c|}{Trans-BDC Block} & \\
 \midrule
 BDC Block & $3\times3_{dw}$ & 3 & - & 208 & 1 &  \\
  & $1\times1_{dw}$ & 1 & - & 208 & 1 &  \\
  & $3\times3_{dw.sep}$ & 3 , 1 & - & 208 & 1 &  \\
  \cline{2-6}
   & \multicolumn{5}{c|}{N=4} & $8 \times 8$ \\
  \midrule
  ViT Block & \multicolumn{5}{c|}{N=4, H=4} & $8 \times 8$ \\
  \midrule
  \midrule
  FMM & \multicolumn{5}{c|}{T=160} & 8$^2$, 16$^2$, 32$^2$, 64$^2$ \\
  \midrule
  GFLOPs & \multicolumn{5}{c|}{-} & 0.6 \\
 
  \bottomrule
    \end{tabular}
    }
\end{table}

\section{Details of Feed-Forward Network} 
\label{sec:fnn_details}
\begin{figure}[h]
  \centering
  \includegraphics[width=6cm]{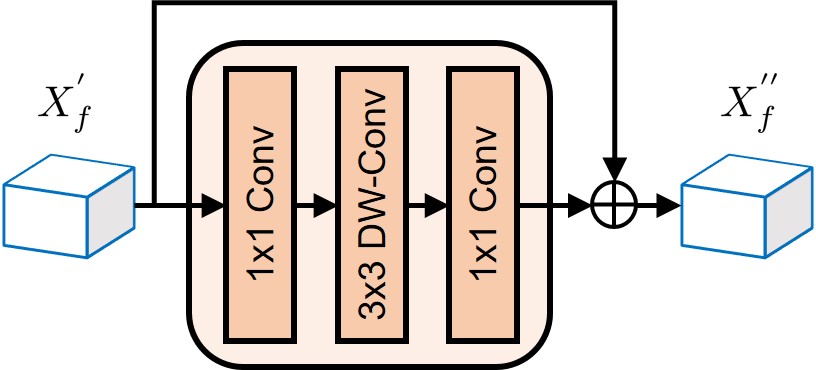}
  \caption{Design of Feed-Forward Network.
  }
  \label{fig:feed_forward_network}
\end{figure}

In the proposed ContextFormer model, the unified output semantics from the BDC block and ViT block are passed through the Feed-Forward Network. For the Feed-Forward Network, we have integrated depth-wise convolution layer between $1 \times 1$ convolution layers and to further minimize the computational complexity, expansion factor of two is incorporated which helps the Trans-BDC block to augment the capturing of global semantics. The design of Feed-Forward Network is shown in~\cref{fig:feed_forward_network}.

\section{Visual Results} 
\label{sec:visual_results}
\begin{figure*}[h]
  \centering
  \includegraphics[width=16.8cm]{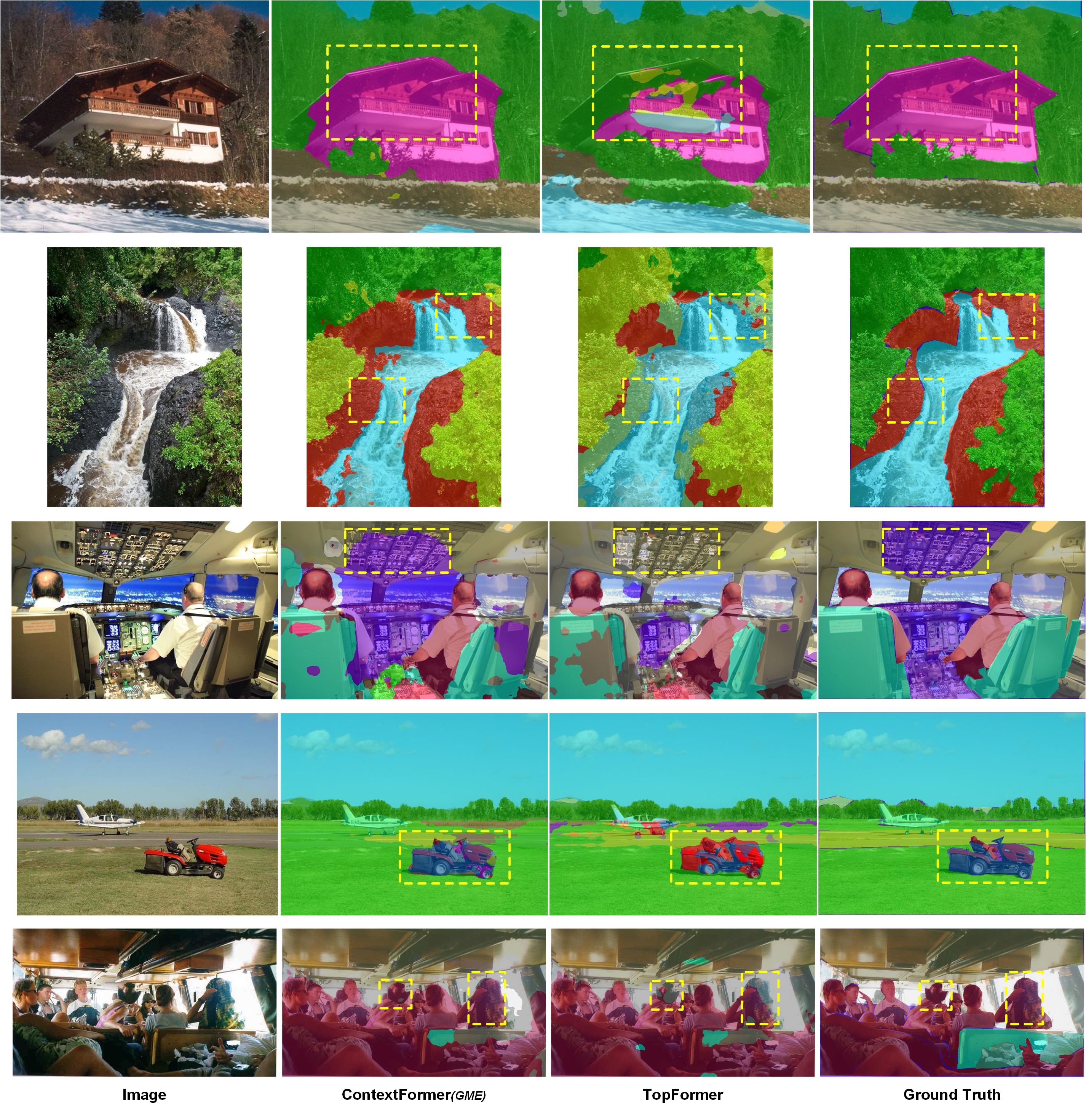}
  \caption{Visual results on the ADE20K validation set. Our proposed model illustrates superior consistency with the ground truth segmentation results, highlighting its robustness. 
  }
  \label{fig:proposed_method_visual_results_supp}
\end{figure*}

~\cref{fig:proposed_method_visual_results_supp} shows additional visual results of the proposed ContextFormer model with original images, ground-truth, TopFormer, and ContextFormer$_{(GME)}$. The proposed model demonstrates superior segmentation performance on the validation set of the ADE20K benchmark dataset, highlighting its robustness. 

\section{Limitations and Future Work} 
\label{sec:limitations_futurework}

The proposed model exhibits strong performance and efficiency in semantic segmentation tasks. However, certain limitations warrant further investigation. A notable constraint for lightweight efficient models, including ours, is their dependency on pre-training with the ImageNet-1K dataset to achieve optimal performance; the absence of such pre-training leads to a significant decline in effectiveness. Future work will focus on improving the model's robustness and efficiency across diverse datasets and practical scenarios. Expanding the scope of testing and validation will be crucial to establish the model's generalizability. Furthermore, emphasis will be placed on adapting and refining the model for real-world applications to enhance its practical utility.